# A General Purpose Inference Engine for Evidential Reasoning Research

*Richard M. Tong, Lee A. Appelbaum, Daniel G. Shapiro*

Advanced Decision Systems
201 San Antonio Circle, Suite 286
Mountain View, CA 94040.

## 1. INTRODUCTION

The purpose of this paper is to report on the most recent developments in our ongoing investigation of the representation and manipulation of uncertainty in automated reasoning systems. In our earlier studies (Tong and Shapiro, 1985) we described a series of experiments with RUBRIC (Tong et al., 1985), a system for full-text document retrieval, that generated some interesting insights into the effects of choosing among a class of scalar valued uncertainty calculi. In order to extend these results we have begun a new series of experiments with a larger class of representations and calculi, and to help perform these experiments we have developed a general purpose inference engine.

The first section of the paper reviews the formal model of retrieval that RUBRIC assumes. The next section describes the extended models of uncertainty that the new engine supports. The third section describes the design and implementation of the engine. The final section reports on the experiments currently being performed.

## 2. THE RETRIEVAL MODEL

To set the context for our work, we suppose that there is a database of documents, denoted by $S$, in which the user is potentially interested. In response to a query, Q, the retrieval system returns a set of retrieved documents, denoted R, that are purported to be *relevant* to that particular query. In general, this set R will only be an approximation to the actual set of relevant documents, denoted $R^*$, contained in the database. The IR problem is thus one of developing a retrieval system that makes $R=R^*$ for all queries. Since this cannot be achieved in practice, we must design systems that maximize the intersection of R and $R^*$ (i.e., maximizing *recall*) while minimizing $R-R^*$ (i.e., maximizing *precision*). In RUBRIC the ability to express retrieval requests in terms of hierarchical collections of rules with attached uncertainty weights results in both higher precision and higher recall than found in more conventional system.

In developing our formal model we start from the view that the primary function of a retrieval system is to select a sub-set of the documents in the database as defined by their *relevance* to the user's query. To understand this, we must distinguish between the subject matter of a document and the utility of the document to the user. So for example, a retrieved document might be about the topic of the user's query, but it might not be useful because the user has seen it before, or because it is a summary of a longer document. Thus relevance includes a notion of usefulness for the task at hand. We have no way of describing this idea of a user goal in the current version of RUBRIC, so we approximate by asserting that a document is relevant if its subject matter is the same as the subject of the user's query.

Our notion of relevance is further modified by the recognition that in most cases the decision as to whether or not a document is about the topic of a query is inherently vague. That is, a document can be about a topic to a certain degree; ranging from not about a topic at all, to definitely about a topic. Notice too, that a document can be about many topics, so in RUBRIC we do not ask the question "What is the document about?" but rather we ask "Is the document about topic X?" RUBRIC's task is thus to decide to what degree the document under consideration is about the topic of the query.

So suppose that the user has a finite set of retrieval concepts, $C$, of interest:

$$C \triangleq \{c_1, c_2, \cdots, c_M\}$$

and that the database, $S$, contains a finite number of documents:

$$S \triangleq \{s_1, s_2, \cdots, s_N\}$$

then we assume that there is an underlying relevance relation, $R$, from $C$ to $S$ such that $R(m,n)$ is the true relevance of document $s_n$ to concept $c_m$. Notice that the range of this relation is left unspecified since we are concerned with a number of possible representations for the degrees of uncertainty. Thus a row-tuple of $R$, denoted $R^{\cdot}(c)$, is the "uncertain" subset of documents that are relevant to concept $c$.

A dual notion which arises is that which we call *typicality*. It refers to the fact that a document is an exemplar, to varying degrees, of documents that are about a particular retrieval concept. Since this notion of typicality also admits of degrees, we can define a fuzzy typicality relation, $T$ from $C$ to $S$ such that $T(m,n)$ is the typicality of document $s_n$ for concept $c_m$. A row-tuple of $T$, denoted $T^{\cdot}(c)$, can then be viewed as defining the subset of typical documents for that concept.

Now if we insist that $R$ and $T$ are in fact one and the same, we see that a rule-based query in RUBRIC can be thought of in two distinct, but related, ways. In the first interpretation, the rules used by RUBRIC constitute an algorithm for determining the degree of membership of a document in the set of relevant documents. That is, they define a network of relevance relationships for deciding how knowledge about the presence of a low-level concepts indicates the presence of a higher-level concept. In the second interpretation, the rules used by RUBRIC act as a definition of what constitutes a typical document. That is, they suggest what low-level concepts should appear if the goal is to find documents that are about a higher-level concept.

## 3. MODELS OF UNCERTAINTY

Since a major goal of the RUBRIC system is to provide a test-bed for the exploration of issues in uncertainty representation, we provide a number of representations that can be selected by the users of RUBRIC. In the first class of models, uncertainty is represented as a scalar in the interval $[0,1]$ and the calculi are based on infinitely-valued logic. In the second class of models, sub-intervals of the $[0,1]$ interval are used to represent uncertainty and the calculi are based on both probabilistic and logical assumptions. In the third class, uncertainty is represented by a linguistic variable whose domain is the unit interval and is manipulated using fuzzy logic. We consider each of these briefly:

### 3.1. Scalar Representations

In our first class of models we assert that relevance can be quantified as a real number in the interval $[0,1]$. Since there is an obvious fuzzy set theoretic interpretation of these values, we draw upon work on many-valued logics (Rescher, 1969), and on the use of triangular-norms as models of fuzzy set intersection (e.g., Dubois and Prade, 1982), to help us construct a calculus of scalar relevance values.

We allow three distinct pairs of definitions for conjunction (the *and* connective), and disjunction (the *or* connective). Negation is defined by $v[\text{not } A] = 1 - v[A]$.

In two-valued logic the *modus ponens* syllogism allows $B$ to be inferred from $A$ and $A => B$. In an infinitely-valued logic, we need to extend this idea so that the relevance of $B$, denoted $v[B]$, can be computed from any given $v[A]$ and $v[A => B]$, where "$=>$" is some infinitely-valued implication. Functions that allow us to compute $v[B]$ are called detachment operators (and are denoted *). It is usual to define them so that for a given definition of $=>$, $v[A] * v[A => B]$ is a lower bound on the value of $v[B]$. We allow four distinct definitions corresponding to four distinct infinitely-valued implication operators.

### 3.2. Interval Representations

Our second class of models assumes that uncertainty can be represented by a sub-interval of the unit interval. That is rather than specify our uncertainty by means of single number we allow it to be expressed by means of a pair of numbers that correspond to the lower and upper values of the interval. So we have an interval such as $[\alpha, \beta]$ in which $\alpha, \beta \leq 1.0$ and $\alpha \leq \beta$. The primitive semantics of these intervals are that $[0,0]$ corresponds to absolutely false (not relevant), $[1,1]$ corresponds to absolutely certain (fully relevant), and $[0,1]$ corresponds to completely unknown. Other sub-intervals reflect greater or lesser certainty, with the width of the interval (i.e. $\beta - \alpha$) representing additional information about the confidence we have in these certainty measures.

The advantage of such a representation is that it allows us to capture finer shades of meaning with respect to the uncertainty associated with propositions. The recent interest in AI in such interval-valued models largely derives from Shafer's influential book (1976) as well as more recent work that attempts to make Shafer's theory practically useful (see for example Gordon and Shortliffe, 1985). However, there are a number of competing models which although using an interval representation start from very different perspectives. We do not have the space to discuss these differences in detail but refer the reader to Tong *et al.* (1986) for a more detailed consideration of the issues.

RUBRIC-III provides specific support for four distinct interval models, although it is trivial to add additional models as required. Specifically, we include:

[1] an interval-valued calculus based on Appelbaum and Ruspini (1985). This is an interval-valued probability calculus which makes minimal assumptions about the relationships between the events being modelled.

[2] a calculus based on the work of Baldwin (1986). This is also an extension of classical probability but with additional assumptions about the relationship between $\alpha$ and $\beta$. In this model $\alpha$ corresponds to the "support" for the base proposition, and $\beta$ is equal to complement with respect to one of the support for the negation of the base proposition. In this respect it is similar to Shafer's model, but it differs in other particulars.

[3] a calculus derived from application of Zadeh's *extension principle* (1973) to the scalar calculi described in section 3.1. This is thus a logical view with minimal assumptions about the the relationships between $\alpha$ and $\beta$.

[4] a calculus derived from a combined application of *modus ponens* and *modus tollens* as discussed by Martin-Clouaire and Prade (1985). This is also a generalization of the calculi described in section 3.1 but with additional assumptions about the logical relationships between the relevant propositions.

### 3.3. Linguistic Representations

The third class of models assumes that uncertainty is primarily linguistic and that it can be operationalized by means of Zadeh's concept of a linguistic variable (Zadeh, 1975) together with a corresponding fuzzy logic. So in this model, uncertainty values are expressed by terms such as "very certain," "more or less false" and "not false but not certain."

RUBRIC-III supports this form of representation by providing the user with the ability to define and calibrate primary linguistic terms from which more complex terms can be generated automatically. The underlying computations are performed using extensions of the logical calculi defined for intervals in section 3.2 so that they can be applied to the fuzzy set definitions of the uncertainty terms.

## 4. THE INFERENCE ENGINE

The RUBRIC inference engine is designed for maximum flexibility. The emphasis has been placed on allowing for multiple uncertainty representations without having to modify any system code. The engine is thus a very useful research tool, although precisely because of its flexibility it is limited in its execution and space efficiency. It is an outgrowth of the earlier work by Appelbaum (Appelbaum and Ruspini, 1985), and makes use of the SOPE object-oriented programming environment developed at ADS (Cation, 1986).

The inference engine operates in two passes. First, a backward inference graph is explicitly created with the goal concept (from the user query) represented by the root node. Second, the inference graph is evaluated by assessing the relevance of the terminal nodes and then propagating and combining relevance values back up the graph to the root. The default operation mode generates the complete backward graph based on the rules, thus one graph may be used to test the relevancy of a collection of documents to a single concept without re-accessing the rule-base.

The implementation of the inference engine is based on the object-oriented programming paradigm. Each node of the inference graph is an object with associated handlers (methods) for expansion and evaluation. Rules are also objects, as are the collections of relevance representations, relevance calculi and the inference-graph itself. The backward and forward evaluation processes are carried out as a series of message passing between the different components mentioned above. The actual details on how each node is represented, expanded and evaluated will be presented at the workshop. The representation and utilization of the different uncertainty calculi objects will also be presented.

Since each node is an independent object with its own expansion and evaluation handlers, assigning different uncertainty functions is a very straight-forward procedure. Each node contains the name of the function to apply (in the LISP sense) to evaluate the relevance of the node given the relevance of its children. There is no restriction on the number of functions which may be employed; each node may use its own function. In practice, however, each node will point to a set of functions which are logically bound together. For example in a fuzzy set interpretation, min, max, and times would be three LISP functions for conjunction, disjunction, and detachment.

Three important "high level" considerations are addressed by this new engine: two control features and an explanation capability. One control feature is the creation of one node for each concept and terminal (a clause which contains only strings to be searched for in the documents) independently of how many times the concept or terminal may appear in the rules. It is from this feature that we produce a graph rather than a tree. It is still necessary, however, to keep track of all the references to the item. This is done through the arcs which link the concept or terminal nodes to their predecessors.

The second control feature is the actual control over graph traversal. Since the complete graph is generated before any evaluation, it is not possible to generate the minimal tree for a given application. However, it is a simple matter to order the rule nodes so that EVIDENCE rules are evaluated before IMPLIES rules and simpler IMPLIES before more complex ones. (EVIDENCE and IMPLIES rules are two of the several rule types available in RUBRIC.) This can be done in the expansion handlers. There is also the possibility of limiting the branches traversed below any node representing a complex clause. For example, in a node representing a conjunction, it may be possible to ascertain that the final result will be 0 (or below some predefined threshold given in the rule base) before all clauses have been evaluated. In this case it is not necessary to proceed any further with that node. There are similar tests for disjunctions and evidence combinations. Of course, these tests are all heavily dependent on the uncertainty calculus employed and thus require the user to supply all the tests (in LISP) before graph expansion.

Finally, a significant degree of explanation is possible through careful use of the action slot of rules. For example, it is possible to write out any string or variable binding to any node of the graph during graph evaluation. This could be simply printing the relevance value of the concept or generate a lengthy description of the environment at the time of evaluation. By traversing up the graph (towards the root) it is possible to reconstruct the reasoning employed in assigning any particular relevance value.

## 5. EXPERIMENTS

There are two types of comparison that we wish to perform with the aid of the RUBRIC inference engine. In the first case, we are concerned with the behavior of variants within the model class. Within each model class the applicability of the representation to the various uncertainty considerations must be addressed. For example, in our first series of experiments we explored the applicability of scalar representations based on multi-valued logics. We noted that certain variants seemed inappropriate and discussed the reasons for this.

In the second case, we are concerned to compare the different model classes with each other. The primary goal here being to look for insights which will allow users of evidential reasoning schemes to select the appropriate model, or models, for the problem at hand. Since the purpose of systems like RUBRIC is to make decisions (i.e., is this document relevant to the topic of the user's query?) if we find that under certain circumstances we would select exactly the same documents whichever model we used then we might rationally choose the easiest technique to represent. Conversely, if we find that under certain circumstances one class of models performs better than another, we need to understand which features of the model lead to the improved performance.

Our most recent series of experiments is a comparison of the four variants of interval model described earlier. As in the earlier series, we have made use of a small collection of documents taken from the Reuters News Service and worked with a number of queries that are concerned with the concept of terrorism. The experimental procedure is also as before; repeated application of the query to the database changing only the uncertainty representation, and then observing the changes in retrieval performance.

The space limitations imposed by the short paper format prevent us from describing the results in any detail, so we will defer a full presentation of them until the workshop itself. Some of the interesting issues that we address are the ease with which users can enter interval values for the rules, interpretation of the interval-valued relevance scores, measures of performance, and the detailed behavior of the interval boundaries (especially the upper bound). We compare the performance of the interval variants, and contrast the results with those from our earlier work.

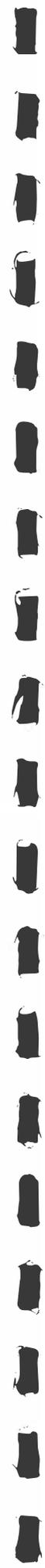